\documentclass[conference]{IEEEtran}
\IEEEoverridecommandlockouts
\usepackage{cite}
\usepackage{amsmath,amssymb,amsfonts}
\usepackage{algorithmic}
\usepackage{graphicx}
\usepackage{textcomp}
\usepackage{xcolor}
\usepackage{booktabs}
\usepackage{multirow}
\def\BibTeX{{\rm B\kern-.05em{\sc i\kern-.025em b}\kern-.08em
    T\kern-.1667em\lower.7ex\hbox{E}\kern-.125emX}}
\begin{document}

\title{Generalizable Collaborative Search-and-Capture in Cluttered Environments via Path-Guided MAPPO and Directional Frontier Allocation\\
}

\author{
    \IEEEauthorblockN{
        Jialin Ying, 
        Zhihao Li, 
        Zicheng Dong, 
        Guohua Wu and 
        Yihuan Liao\IEEEauthorrefmark{1}
    }
    \\ \vspace{1ex}
    \IEEEauthorblockA{
        Department of Automation, Central South University, Changsha, China
    }
    \thanks{*Corresponding author: Yihuan Liao (email: 225189@csu.edu.cn)}
}

\maketitle

\begin{abstract}
Collaborative pursuit-evasion in cluttered environments presents significant challenges due to sparse rewards and constrained Fields of View (FOV). Standard Multi-Agent Reinforcement Learning (MARL) often suffers from inefficient exploration and fails to scale to large scenarios. We propose \textbf{PGF-MAPPO} (Path-Guided Frontier MAPPO), a hierarchical framework bridging topological planning with reactive control. To resolve local minima and sparse rewards, we integrate an A*-based potential field for dense reward shaping. Furthermore, we introduce \textbf{Directional Frontier Allocation}, combining Farthest Point Sampling (FPS) with geometric angle suppression to enforce spatial dispersion and accelerate coverage. The architecture employs a parameter-shared decentralized critic, maintaining $O(1)$ model complexity suitable for robotic swarms. Experiments demonstrate that PGF-MAPPO achieves superior capture efficiency against faster evaders. Policies trained on $10 \times 10$ maps exhibit robust \textbf{zero-shot generalization} to unseen $20 \times 20$ environments, significantly outperforming rule-based and learning-based baselines.
\end{abstract}

\begin{IEEEkeywords}
Multi-Robot Systems, Deep Learning in Robotics and Automation, Path Planning for Multiple Mobile Robots or Agents, Distributed Robot Systems, Pursuit-Evasion Games.
\end{IEEEkeywords}

\section{INTRODUCTION}

Multi-robot collaborative pursuit-evasion (search-and-capture) is fundamental in robotics, with applications from search-and-rescue to perimeter security \cite{ref_pomdp_pursuit}. Unlike coverage tasks, this requires robots to efficiently explore unknown environments with static obstacles while coordinating to trap a dynamic evader under \textit{limited Field of View (FOV)} and sparse sensor information. Traditional approaches rely on geometric decompositions like Voronoi partitions \cite{ref_voronoi}, assuming perfect global knowledge, limiting adaptability. Multi-Agent Reinforcement Learning (MARL), particularly MAPPO \cite{ref_mappo}, shows promise but struggles in large-scale search tasks due to \textit{sparse rewards}. Without explicit topological guidance, RL agents engage in "blind exploration," getting trapped in local minima or clustering. Large-scale swarm deployment imposes strict computational constraints.

To address these challenges, we propose \textbf{PGF-MAPPO} (Path-Guided Frontier MAPPO), decoupling long-horizon exploration from short-horizon control. We integrate an A*-based path planner into the RL loop, providing dense, topology-aware reward shaping and observation guidance. We design \textbf{Directional Frontier Allocation} combining Farthest Point Sampling (FPS) with geometric angle suppression to generate dispersed sub-goals, accelerating map coverage and preventing the "herd effect." We implement the policy using a parameter-shared, decentralized critic architecture, maintaining constant model complexity ($O(1)$) regardless of team size.

The main contributions are: (1) PGF-MAPPO, a hierarchical framework incorporating topological path guidance and curriculum learning to solve the sparse reward problem in cluttered environments with limited FOV; (2) a novel target assignment strategy using Directional FPS and Hungarian matching with angle suppression, significantly reducing inter-agent conflict and improving search efficiency; (3) extensive experiments demonstrating superior performance over baselines, with policies trained on $10\times10$ maps exhibiting strong \textbf{zero-shot generalization} to larger $15\times15$ and $20\times20$ environments.

\begin{figure*}[t]
    \centering
    \includegraphics[width=1.0\linewidth,trim=0cm 21cm 0cm 0cm, clip]{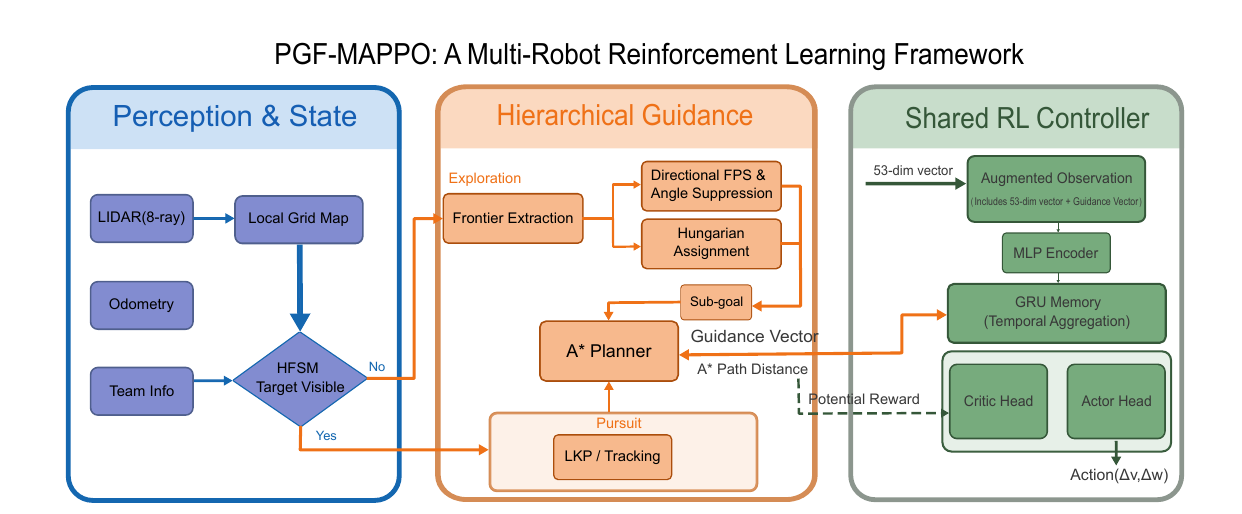} 
    \caption{The proposed PGF-MAPPO framework architecture. The system operates hierarchically: (Left) The \textbf{Perception \& State Module} builds a local belief map and manages mode switching via an HFSM. (Middle) The \textbf{Hierarchical Guidance Module} generates high-level sub-goals using Directional Frontier Allocation (during exploration) or LKP tracking (during pursuit), and computes an A* path to provide topological guidance. (Right) The \textbf{Shared RL Controller} processes the augmented observation—including the A* guidance vector—to output continuous kinematic actions. Dense rewards derived from the A* potential field facilitate stable training.}
    \label{fig:architecture}
\end{figure*}

\section{RELATED WORK}

\subsection{Multi-Robot Pursuit and Learning-Based Control}
Traditional geometric approaches like Voronoi tessellations \cite{ref_voronoi} provide theoretical guarantees but assume holonomic kinematics and perfect global information. Multi-Agent Reinforcement Learning (MARL), particularly MADDPG \cite{ref_maddpg} and MAPPO \cite{ref_mappo}, addresses continuous control but struggles with sparse rewards. Limited Field-of-View (FOV) transforms pursuit into a Partially Observable Markov Decision Process (POMDP) \cite{ref_pomdp_pursuit}, necessitating memory-based architectures \cite{ref_drqn}. IPPO with parameter sharing maintains constant model complexity \cite{ref_ippo}, which we leverage.

\subsection{Coordinated Exploration and Frontier Allocation}
Frontier-based exploration \cite{ref_yamauchi} directs robots to boundaries between known and unknown space. Coordination strategies range from market-based auctions \cite{ref_auction} to Graph Neural Networks (GNN) \cite{ref_gnn_exploration}, but GNN-based methods impose significant overhead while greedy strategies cause the "herd effect." We introduce lightweight Directional Frontier Allocation combining Farthest Point Sampling (FPS) with geometric angle suppression.

\subsection{Hybrid Navigation and Reward Shaping}
Integrating classical planners with RL improves long-horizon navigation \cite{ref_prm_rl}. Hybrid systems require dense rewards that preserve optimal policies. Intrinsic motivation methods like VIME \cite{ref_geodesic} and RND \cite{burda2018exploration} encourage exploration but lack goal directionality. Our framework leverages Potential-based Reward Shaping \cite{ng1999policy}, constructing a potential field based on dynamic A* pathfinding.

\section{METHODOLOGY}

\subsection{Problem Formulation}
We formulate the multi-robot pursuit-evasion task as a Decentralized Partially Observable Markov Decision Process (Dec-POMDP), defined by the tuple $\langle \mathcal{N}, \mathcal{S}, \mathcal{A}, \mathcal{O}, \mathcal{P}, \mathcal{R}, \gamma \rangle$.
The team consists of $N=3$ pursuers, denoted by $\mathcal{N}=\{1, \dots, N\}$. 
At time step $t$, the global state $s_t \in \mathcal{S}$ describes the full environment status but is hidden from agents.
Each agent $i$ receives partial observation $o_{i,t} \in \mathcal{O}$ and executes action $a_{i,t} \in \mathcal{A}$ based on local policy $\pi_{\theta}(a_{i,t} | o_{i,t})$. The environment transitions according to $\mathcal{P}(s_{t+1} | s_t, \mathbf{a}_t)$ and yields reward $r_{i,t} \in \mathcal{R}$. The objective is to maximize the expected discounted return $J(\theta) = \mathbb{E}[\sum_{t=0}^T \gamma^t r_{i,t}]$, where $\gamma=0.99$.

\subsubsection{Observation Space}
The observation vector $o_{i,t} \in \mathbb{R}^{53}$ consists of: (1) \textbf{Proprioception} ($\mathbf{o}_{kin}$): normalized position $(x, y)$, ego-centric velocity $(v_x, v_y)$, and heading $(\cos \theta, \sin \theta)$; (2) \textbf{Exteroception} ($\mathbf{o}_{sense}$): 8-ray wall LiDAR and 8-ray teammate LiDAR readings; (3) \textbf{Navigation \& Memory} ($\mathbf{o}_{nav}$): the \textbf{A* guidance vector} $\mathbf{v}_{guide}$, the Last Known Position (LKP) of the target, and a $3\times3$ local occupancy patch from the agent's internal belief map; (4) \textbf{Identification} ($\mathbf{o}_{id}$): a one-hot vector encoding the agent's unique ID.

\subsubsection{Action Space and Dynamics}
We employ a continuous action space controlling differential drive kinematics. The action $a_{i,t} = [\Delta \theta_t, \Delta v_t] \in [-0.6, 0.6] \times [-0.4, 0.4]$ represents incremental change in heading and linear speed:
\begin{equation}
    \begin{aligned}
    \theta_{t+1} &= \theta_t + \Delta \theta_t, \\
    v_{t+1} &= \text{clip}(v_t + \Delta v_t, 0, v_{max}),
    \end{aligned}
\end{equation}
where $v_{max}=1.2$ m/s. A physics engine handles collisions with damping and wall-sliding, and a repulsive force activates when inter-agent distance is below $0.6$ m.

\subsection{Policy Architecture: The PGF-MAPPO Framework}
\label{sec:policy}
We propose \textbf{PGF-MAPPO} (Path-Guided Frontier Multi-Agent PPO) for collaborative pursuit. While grounded in Multi-Agent PPO, we optimize the architecture for scalability and deployment on \textbf{resource-constrained edge devices} (e.g., embedded UAV processors).

\subsubsection{Decentralized Implementation with Parameter Sharing}
PGF-MAPPO adopts a \textbf{parameter-shared, decentralized critic} structure. Our network model contains approximately \textbf{1.87 million} trainable parameters, occupying $\approx$ \textbf{7.5 MB} of memory per instance. For $N=10$ agents, a non-shared architecture would require $18.7$ million parameters ($\approx 75$ MB). By sharing weights $\theta$ across all $N$ pursuers, our method maintains \textbf{constant complexity $O(1)$}, with total memory footprint locked at 7.5 MB regardless of swarm size.

\subsubsection{Network Structure}
The shared policy network processes the 53-dimensional observation $o_{i,t}$ through: (1) a 2-layer MLP (512 hidden units, LayerNorm, Tanh activation) extracting features; (2) a GRU (hidden state size 512) aggregating historical context to resolve partial observability; (3) dual heads: Actor Head outputting mean $\mu$ and log-std $\log \sigma$ for Gaussian action distribution, and Critic Head outputting value estimate $V(o_{i,t})$ based on local observations.

\subsection{Topological Path Guidance via A*}
\label{sec:guidance}

Standard RL exploration fails in cluttered environments due to local minima and sparse rewards. We integrate an A* planner leveraging the agent's internal belief map to provide topological guidance, functioning as both augmented observation and dense reward signal.

\subsubsection{Path Generation and Vector Observation}
At every $k$-th time step (default $k=3$), the planner computes a collision-free path $\tau^* = \{w_0, w_1, \dots, w_M\}$ from the agent's current position $p_t$ to the current sub-goal $g_t$. We extract a \textbf{local waypoint} $w^*$ as the furthest point on the path within a lookahead distance $d_{lookahead}=1.5$ m:
\begin{equation}
    w^* = \operatorname*{arg\,max}_{w \in \tau^*, \, \|w - p_t\| \le d_{lookahead}} \|w - p_t\|.
\end{equation}
The \textbf{guidance vector} $\mathbf{v}_{guide} \in \mathbb{R}^2$ is computed by transforming the direction to $w^*$ into the agent's body frame:
\begin{equation}
    \mathbf{v}_{guide} = R(-\theta_t) \cdot \frac{w^* - p_t}{\|w^* - p_t\|},
\end{equation}
where $R(-\theta_t)$ is the rotation matrix by the agent's current heading. This vector feeds into the actor network.

\subsubsection{Topological Potential-based Reward}
We use the A* path length to construct a dense reward signal. Unlike Euclidean distance, the A* distance $D_{A^*}(p_t, g_t)$ represents the true geodesic cost in the environment. Following Potential-based Reward Shaping \cite{ng1999policy}, we define the potential function $\Phi(s_t) = -D_{A^*}(p_t, g_t)$ and the shaping reward:
\begin{equation}
    R_{pot} = \lambda_{pot} \cdot (\gamma \Phi(s_{t+1}) - \Phi(s_t)),
\end{equation}
where $\gamma$ is the discount factor and $\lambda_{pot}$ is a scaling coefficient. This formulation ensures the auxiliary reward aids convergence without altering the optimal policy.

\subsection{Hierarchical Coordination and Frontier Allocation}
\label{sec:allocation}
We implement a \textbf{Hierarchical Finite State Machine (HFSM)} coordinating search-pursuit transitions. Each agent toggles between \textbf{Pursuit Mode} (target visible or valid LKP exists) and \textbf{Exploration Mode} (lacks target information). Within Exploration Mode, agents operate a sub-FSM: (1) \textbf{Assignment \& Locking}: receives sub-goal $g^*$ from the central allocation algorithm (Sec. \ref{sec:allocation_algo}), locked until termination; (2) \textbf{Approach}: navigates towards $g^*$ using A* guidance, transitioning to Sweep phase once within proximity threshold ($d_{approach} < 2.0$ m); (3) \textbf{Local Sweep}: switches to local \textbf{Greedy Coverage}, clearing adjacent unexplored grid cells until fully mapped or maximum step limit is reached, then unlocks and requests a new assignment.

\subsubsection{Directional Frontier Assignment Algorithm}
\label{sec:allocation_algo}
When an agent requests a new assignment, we employ a spatially dispersed strategy. We generate $K$ candidate sub-goals $\mathcal{G}$ via \textbf{Directional Farthest Point Sampling (FPS)} with \textbf{Sector Suppression} (threshold $\phi_{suppress}=45^\circ$) to ensure angular diversity. The optimal assignment is solved via the Hungarian Algorithm using a kinematic-aware cost matrix:
\begin{equation}
    C_{i,m} = \frac{\|p_i - g_m\|_2}{v_{max}} + w_{angle} \cdot |\Delta \theta(p_i, g_m)|,
\end{equation}
where $w_{angle}=2.0$ penalizes targets requiring large turns.

\subsection{Reward Function Design and Curriculum}
\label{sec:reward}
\subsubsection{Composite Reward Structure}
The total reward $r_t^i$ for agent $i$ at time step $t$ is a weighted sum of four components:
\begin{equation}
    r_t^i = r_{mission} + r_{safety} + r_{guide} + r_{explore}.
\end{equation}

\textbf{1) Mission Rewards ($r_{mission}$):}
We apply constant time penalty $\lambda_{time}$ per step to encourage rapid capture. Upon successful capture (distance to target $< d_{cap}$), agents receive large sparse bonus $C_{cap}$. We distinguish "Clean Capture" (no collisions) from standard capture, awarding higher bonus ($+6000$) for the former.

\textbf{2) Safety Penalties ($r_{safety}$):}
We impose penalties for collisions with obstacles or teammates ($C_{col}$) and for keeping static ($C_{static}$) to enforce physical constraints:
\begin{equation}
    r_{safety} = - C_{col} \cdot \mathbb{I}_{collision} - C_{static} \cdot \mathbb{I}_{v \approx 0}.
\end{equation}

\textbf{3) Guidance Rewards ($r_{guide}$):}
Includes \textbf{Potential-based Reward} $R_{pot}$ (Sec. \ref{sec:guidance}, Eq. 6) derived from A* path deltas, and \textbf{Alignment Reward} encouraging velocity vector $\mathbf{v}_t$ to align with A* guidance vector $\mathbf{v}_{guide}$:
\begin{equation}
    r_{align} = \lambda_{align} \cdot \text{clip}(\mathbf{v}_t \cdot \mathbf{v}_{guide}, 0, 1) \cdot \|\mathbf{v}_t\|.
\end{equation}

\textbf{4) Exploration Rewards ($r_{explore}$):}
Active only in Exploration Mode, this intrinsic reward incentivizes map coverage by rewarding visits to new grid cells and reducing distance to assigned frontier sub-goal.

\begin{figure*}[t]
    \centering
    \includegraphics[width=1.0\linewidth]{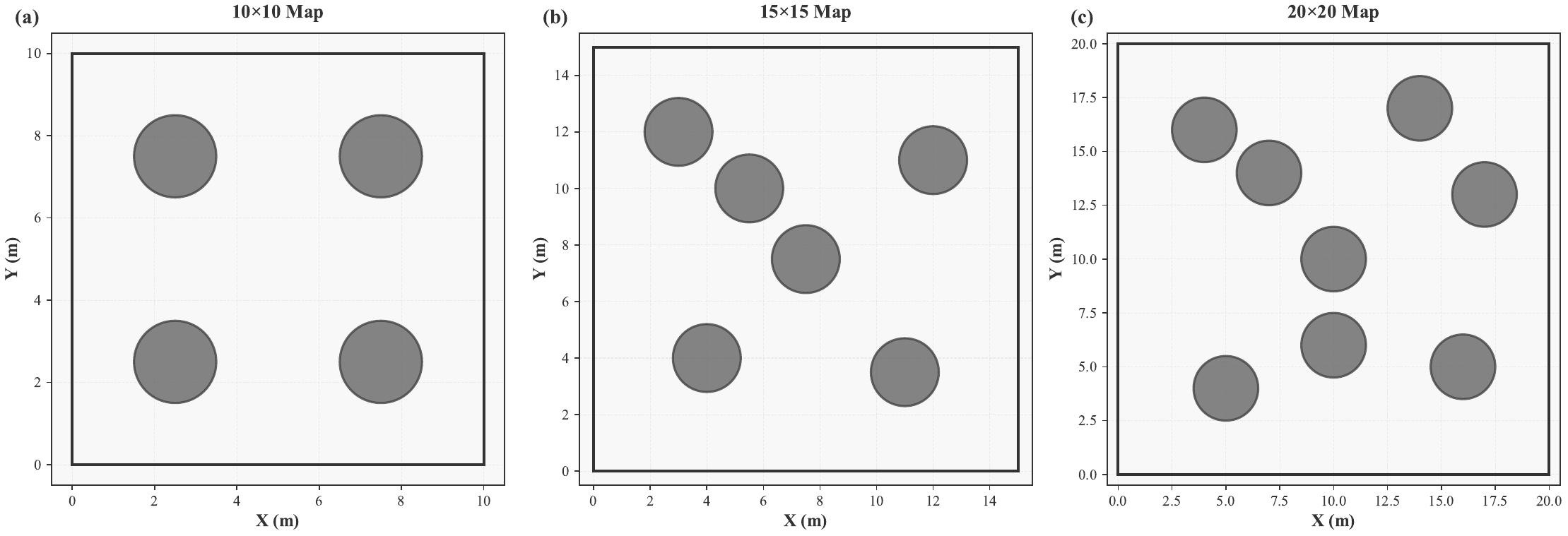} 
    \caption{Evolution of environmental complexity. \textbf{(a) Training Environment ($10\times10$)}: A fixed layout with sparse obstacles used for all curriculum stages. \textbf{(b) \& (c) Evaluation Environments}: Unseen $15\times15$ and $20\times20$ maps with randomly generated, dense obstacles. The transition from (a) to (c) represents a significant leap in difficulty, testing the zero-shot generalization capability of the PGF-MAPPO policy.}
    \label{fig:map_setup}
\end{figure*}

\subsubsection{Curriculum Scheduling}
We employ a 5-stage curriculum: \textbf{Stage 1-2 (Warm-up)}: high observability ($R_{fov}=10 \to 4$ m) and lenient penalties ($\lambda_{time}=-0.5, C_{col}=-10$) to learn basic locomotion and target tracking; \textbf{Stage 3 (Information Bottleneck)}: FOV minimized to $1.5$ m with boosted exploration weights, forcing agents to rely on A* guidance and Frontier Allocation; \textbf{Stage 4-5 (Hard Constraints)}: target speed increases ($1.1 \to 1.3$ m/s) exceeding agents' speed, with severe penalties ($\lambda_{time}=-2.0, C_{col}=-20$) requiring precise, time-optimal coordination.

\section{EXPERIMENTS}

\subsection{Experimental Setup}

\subsubsection{Simulation Environment}
The experimental environments are visualized in Fig. \ref{fig:map_setup}. Training utilized a fixed $10\times10$ m map containing four circular obstacles. Evaluation extended to $15\times15$ m and $20\times20$ m scales with randomly generated obstacles, never seen during training. Simulation operates at time step $\Delta t = 0.2$ s with maximum episode length $T_{max}=512$ steps.

\subsubsection{Agent Dynamics}
Both pursuers and the evader are modeled as non-holonomic agents with physical radius $r_{agent}=0.25$ m. The state of each agent $i$ at time $t$ is $s_t^i = [x_t, y_t, \theta_t, v_t]$. The action space consists of steering and acceleration commands $a_t^i = [\Delta \theta_t, \Delta v_t]$, constrained by $\Delta \theta \in [-0.6, 0.6]$ rad and $\Delta v \in [-0.4, 0.4]$ m/s. The kinematic update follows:
\begin{equation}
\begin{cases}
v_{t+1} = \text{clip}(v_t + \Delta v_t, 0, v_{max}) \\
\theta_{t+1} = \theta_t + \Delta \theta_t \\
x_{t+1} = x_t + v_{t+1} \cos(\theta_{t+1}) \Delta t \\
y_{t+1} = y_t + v_{t+1} \sin(\theta_{t+1}) \Delta t
\end{cases}
\end{equation}
where $v_{max}=1.2$ m/s. A physics engine handles collisions: velocity is damped (0.2 or 0.9 depending on impact angle) when an agent hits an obstacle, and wall-sliding prevents penetration. A repulsive force activates when inter-agent distance is below $0.6$ m.

\subsubsection{Sensors and FOV}
Agents perceive through 8-ray lidar (uniformly spaced by $45^\circ$) detecting static obstacles and teammates within max range 10 m, and visual target detection constrained to a circular sector with FOV radius $R_{fov}=2.0$ m with occlusion effects.

\subsubsection{Task Definition}
The objective is to capture a single high-speed evader using $N=3$ pursuers. A capture is successful if $||p_i - p_e||_2 < d_{cap}$, where $d_{cap}=0.8$ m.

\begin{figure*}[t]
    \centering
    \includegraphics[width=0.95\textwidth]{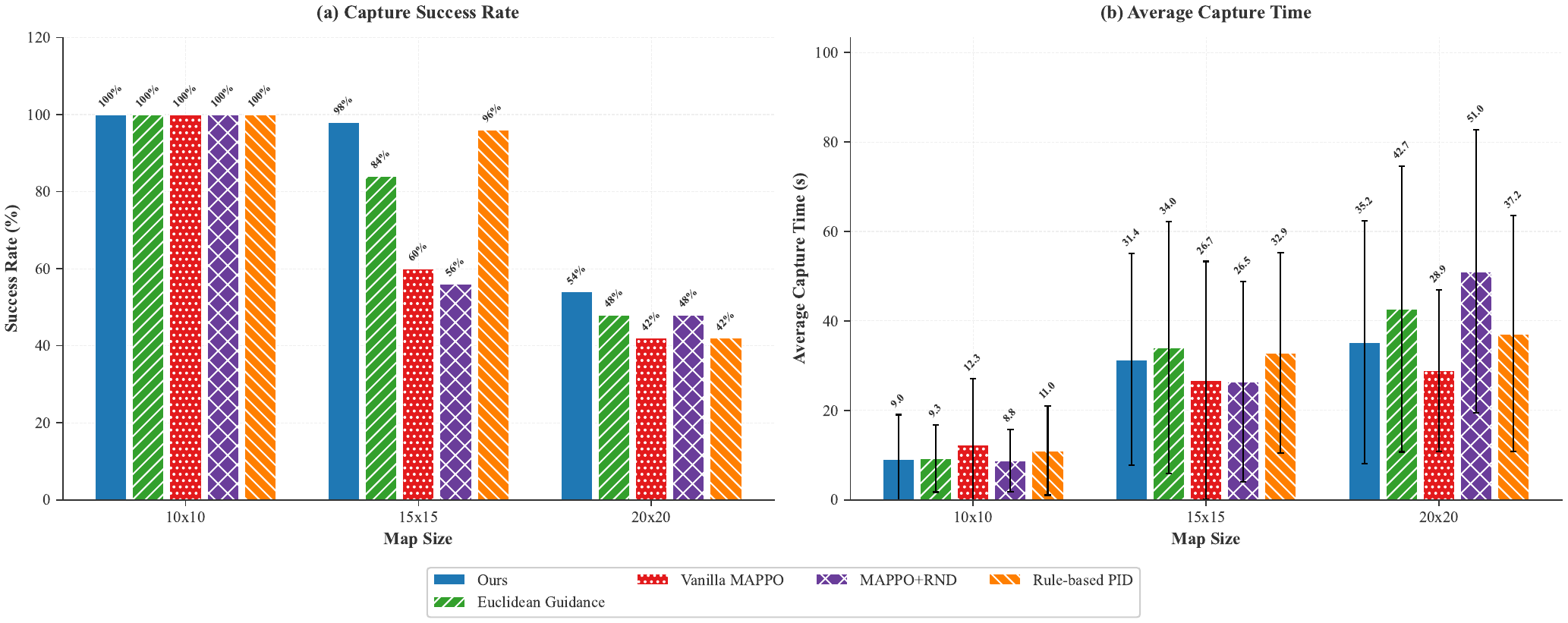}
    \caption{Quantitative comparison results. (a) Capture Success Rate... (b) Average Capture Time...}
    \label{fig:main_results}
\end{figure*}

\begin{table*}[t]
\centering
\caption{Quantitative Comparison of Success Rate (SR) and Average Capture Steps (Steps) Across Different Map Scales}
\label{tab:main_results}
\setlength{\tabcolsep}{4pt}
\begin{tabular}{l|cc|cc|cc}
\toprule
\multirow{2}{*}{\textbf{Method}} & \multicolumn{2}{c|}{\textbf{10$\times$10 (Easy)}} & \multicolumn{2}{c|}{\textbf{15$\times$15 (Medium)}} & \multicolumn{2}{c}{\textbf{20$\times$20 (Hard)}} \\ \cline{2-7} 
 & \textbf{SR} $\uparrow$ & \textbf{Steps} $\downarrow$ & \textbf{SR} $\uparrow$ & \textbf{Steps} $\downarrow$ & \textbf{SR} $\uparrow$ & \textbf{Steps} $\downarrow$ \\ \midrule
PID (Rule) & 100\% & 55 & 96\% & 164 & 42\% & 186 \\
Vanilla MAPPO & 100\% & 62 & 60\% & \textbf{134}* & 42\% & \textbf{144}* \\
MAPPO+RND & 100\% & \textbf{44} & 56\% & 132 & 48\% & 255 \\
Euclidean Guidance& 100\% & 46 & 84\% & 170 & 48\% & 213 \\ \midrule
\textbf{Ours (PGF-MAPPO)} & \textbf{100\%} & 45 & \textbf{98\%} & 157 & \textbf{54\%} & 176 \\ \bottomrule
\end{tabular}
\vspace{1mm}
\footnotesize{\\ * Note: Lower steps in Vanilla MAPPO are due to high failure rates in difficult episodes (survivorship bias).}
\vspace{-3mm}
\end{table*}

\subsubsection{Baselines}
We compare PGF-MAPPO against four baselines: (1) \textbf{Rule-based PID}: sharing the same high-level strategy (Directional Frontier Allocation and A* pathfinding) but using PID control for low-level execution; (2) \textbf{Vanilla MAPPO}: the same parameter-shared architecture but trained without topological path guidance or frontier allocation; (3) \textbf{MAPPO + RND}: Vanilla MAPPO incorporating Random Network Distillation (RND) intrinsic rewards; (4) \textbf{Euclidean Guidance}: replacing A*-based metrics with Euclidean counterparts (potential-based reward uses straight-line distance, guidance vector points directly to target/sub-goal ignoring obstacles).

\subsection{Comparative Performance Analysis}
Quantitative results across three map scales are summarized in Table \ref{tab:main_results} and Fig. \ref{fig:main_results}. In medium-scale environments ($15\times15$), both PGF-MAPPO (98\% SR) and Rule-based PID (96\% SR) achieve near-perfect success rates, while Vanilla MAPPO drops to 60\% SR, validating \textbf{Directional Frontier Allocation}.

\begin{figure*}[t]
    \centering
    \includegraphics[width=0.98\linewidth]{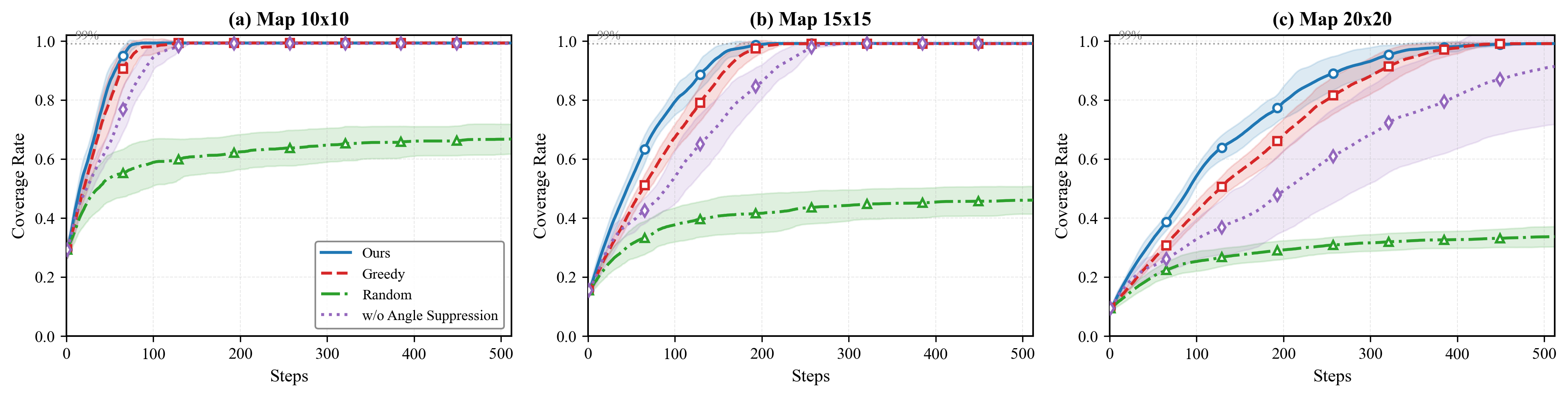} 
    \caption{Map coverage rate over time steps across three scales. (b) and (c) show that our strategy (solid blue line) explores the environment significantly faster than baselines. The comparison with "w/o Angle Suppression" (purple dotted line) highlights the critical role of our spatial dispersion mechanism in large-scale maps.}
    \label{fig:coverage}
\end{figure*}

\begin{figure*}[t]
    \centering
    \includegraphics[width=0.98\linewidth]{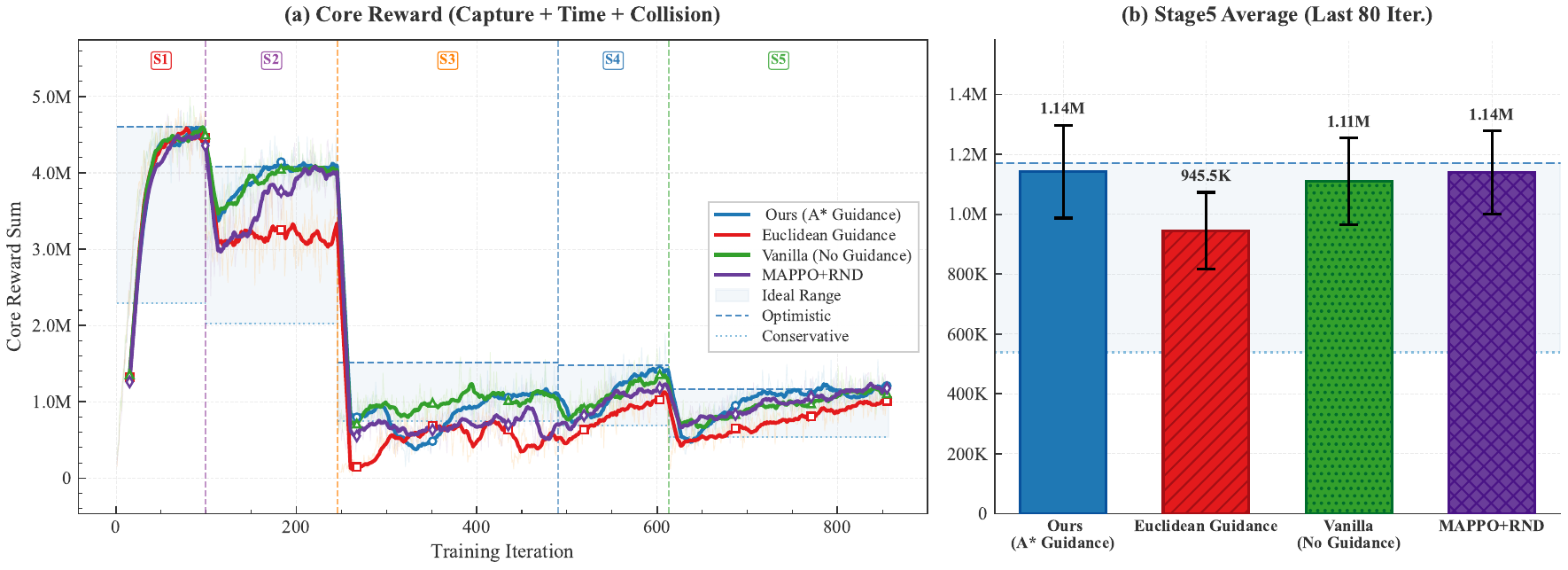} 
    \caption{Training curves of the core reward sum across curriculum stages (S1-S5). While all methods perform similarly in simple stages (S1-S2), the Euclidean Guidance baseline (Red) suffers a performance drop in complex stages (S4-S5). In contrast, our A* Guidance (Blue) maintains a stable upward trend, demonstrating robust convergence in cluttered environments.}
    \label{fig:training_curve}
\end{figure*}
 
\begin{figure}[t]
    \centering
    \includegraphics[width=1.0\linewidth]{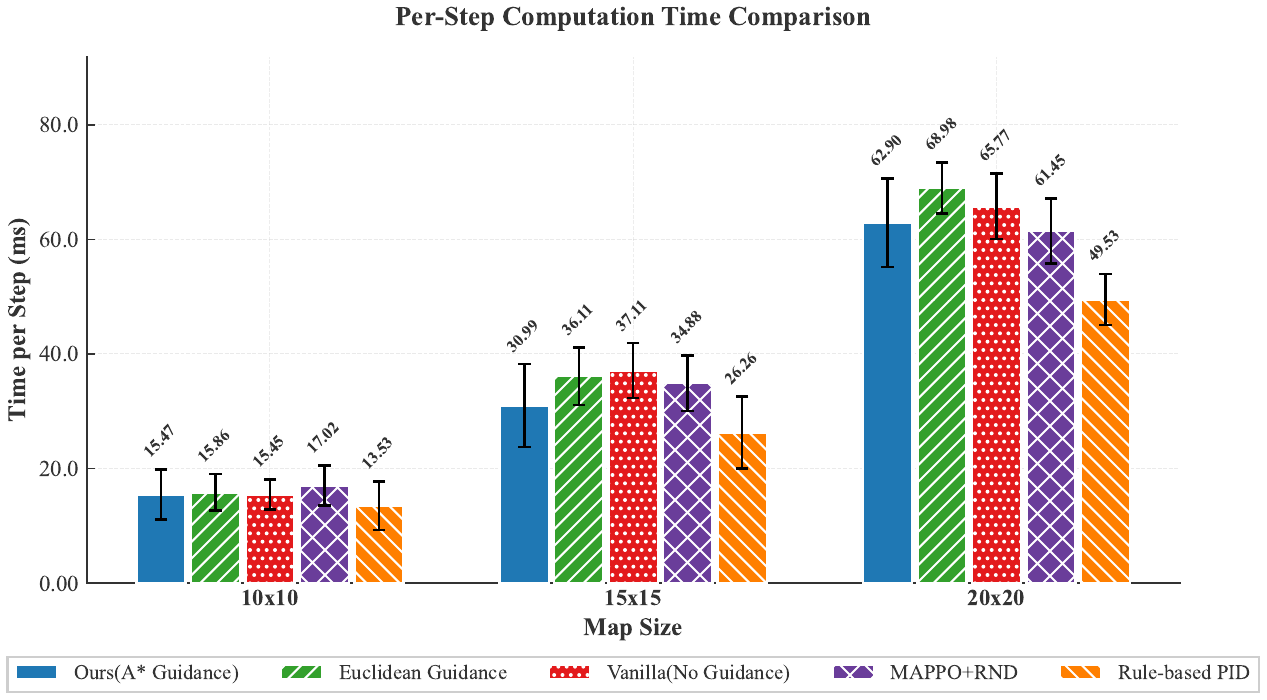} 
    \caption{Per-step computation time comparison. Our PGF-MAPPO (Blue) maintains high efficiency, outperforming Vanilla and Euclidean baselines in medium and large maps. This is attributed to our event-triggered allocation logic and optimized path caching, which minimize redundant calculations.}
    \label{fig:compute_time}
\end{figure}

\begin{figure*}[t]
    \centering
    \includegraphics[width=1.0\linewidth,trim=0cm 0cm 0cm 0cm, clip]{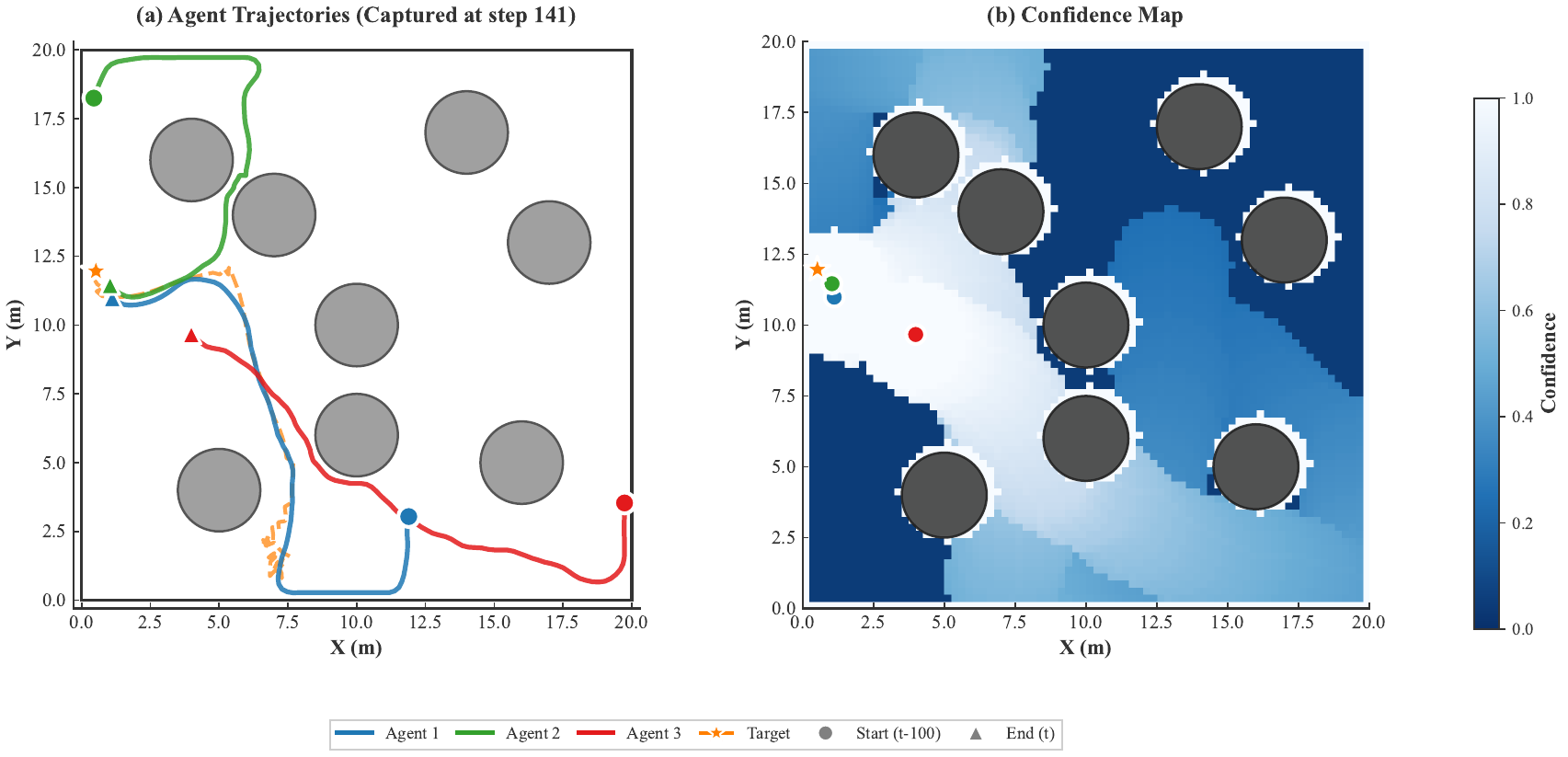} 
    \caption{Qualitative visualization of a successful capture in the unseen $20\times20$ map. \textbf{(a) Agent Trajectories}: The agents (start: circles, end: triangles) successfully corner the dynamic target (star). Note the distinct approach angles, demonstrating the efficacy of our Angle Suppression mechanism. \textbf{(b) Confidence Map}: A snapshot of the internal belief state at the capture moment. The large unexplored regions (dark blue) confirm that once the target is detected (switching to Pursuit Mode), the agents efficiently prioritize interception over exhaustive coverage.}
    \label{fig:qualitative}
\end{figure*}

\subsubsection{Superiority of Learned Control (Ours vs. PID)}
In challenging $20\times20$ maps, the evader's maximum speed ($1.3$ m/s) exceeds pursuers ($1.2$ m/s). Both Rule-based PID and Vanilla MAPPO achieve 42\% success rate, with successes largely stemming from \textit{favorable initializations}. PGF-MAPPO achieves \textbf{54\%} (+12\% margin), indicating RL agents learned \textit{implicit cooperative behaviors} like interception or encirclement, allowing them to trap faster targets even when simple pursuit fails.

\subsubsection{Impact of Topological Guidance (Ours vs. Euclidean)}
In the $15\times15$ map, replacing A* guidance with Euclidean distance reduces the success rate from 98\% to 84\%, confirming topological path awareness is essential. When the direct line of sight to the sub-goal is blocked, Euclidean guidance misleadingly encourages agents to move towards the obstacle surface, while A* guidance provides a valid gradient.

\subsubsection{Capture Efficiency}
Vanilla MAPPO's low step count (134 steps) in $15\times15$ maps is biased. A fairer comparison is between PGF-MAPPO (157 steps) and Rule-based PID (164 steps), which have similar success rates, showing our method trims capture time by about 4.3\%.

\subsection{Generalization and Search Efficiency}
Policies trained on $10\times10$ maps transfer to $15\times15$ and $20\times20$ environments without fine-tuning. In the largest map, the Random baseline stalls below 40\% coverage; Greedy needs $\approx$260 steps to reach 80\%, while PGF-MAPPO needs $\approx$200. Removing angle suppression sharply degrades this gain, confirming directional dispersion is the key to zero-shot generalization.

\subsection{Training Process Validation}
As observed in Fig. \ref{fig:training_curve}, all methods successfully converge to the "Ideal Range" by the end of training, explaining the 100\% success rate on the training map ($10\times10$). Since all models reach the performance ceiling in the training environment, the significant performance gap observed in the $15\times15$ and $20\times20$ test maps is strictly attributable to the \textbf{generalization capability} of the learned policies.

\subsection{Computational Efficiency Analysis}
We evaluated per-step time (policy + environment) over 50 episodes; Fig. \ref{fig:compute_time} shows PGF-MAPPO faster than Vanilla and Euclidean (e.g., 30.99 vs 37.11 ms in $15\times15$) due to: (1) event-triggered Hungarian allocation, (2) A* path caching (refresh every $k=3$), (3) smoother physics. The 2-layer MLP + GRU is lightweight; in the $20\times20$ map, loop time averages 62.90 ms ($\approx 16$ Hz), meeting real-time needs.

\subsection{Qualitative Analysis of Emergent Tactics}
Fig. \ref{fig:qualitative} shows a $20\times20$ episode: agents diverge to form a triangular encirclement, and the confidence map still contains large unexplored areas, indicating a rapid switch from exploration to pursuit once the target is seen.

\section{DISCUSSION}
Our results challenge the prevailing trend of using end-to-end RL for long-horizon navigation. While pure RL (Vanilla MAPPO) theoretically has the capacity to learn navigation policies, it struggles with sparse rewards in large-scale combinatorial spaces. By integrating A* pathfinding as \textit{informational guidance} rather than rigid control, PGF-MAPPO combines the best of both worlds: the planner provides topological awareness, while the RL policy handles fine-grained kinematic control and dynamic obstacle avoidance. This hybrid approach is proven essential for \textbf{zero-shot generalization}, as the "obstacle-aware" navigation logic remains valid regardless of map scale.

A notable finding is the system's ability to capture a target faster than the pursuers ($1.3$ m/s vs. $1.2$ m/s). Since a direct tail-chase is mathematically impossible, the high success rate (54\%) in Stage 5 implies the emergence of implicit coordination. The combination of Directional Frontier Allocation and the shared PPO policy naturally drives agents to cut off escape routes.

\subsection{Limitations and Future Work}
Several limitations exist: (1) \textbf{Dependence on Map Accuracy}: The A* guidance relies on the agent's internal belief map. In highly dynamic environments, the planned path might become obsolete. Future work could explore D* Lite or local reactive planners. (2) \textbf{Communication for Allocation}: While policy execution is decentralized, Frontier Allocation employs the Hungarian algorithm, which implies a centralized computation step. For extremely large swarms ($N > 100$), this could become a communication bottleneck. We plan to investigate fully distributed auction-based mechanisms (e.g., CBBA). (3) \textbf{Sim-to-Real Gap}: Real-world sensor noise and drift are more complex. Validating the PGF-MAPPO framework on physical micro-UAVs is our immediate next step.

\section{CONCLUSION}

In this work, we presented \textbf{PGF-MAPPO}, a hierarchical framework for robust multi-robot pursuit-evasion in cluttered, partially observable environments. By integrating topological A* guidance with a directional frontier allocation strategy, we bridged the gap between long-horizon exploration and reactive kinematic control, solving the sparse reward challenge. Extensive experiments demonstrate superior performance over baselines and remarkable \textbf{zero-shot generalization}, successfully transferring policies from $10\times10$ training environments to complex $20\times20$ scenarios. The proposed parameter-shared architecture maintains constant model complexity ($O(1)$) with minimal computational overhead, enabling scalable, real-time deployment on resource-constrained robotic swarms.

\bibliographystyle{IEEEtran}

\bibliography{reference} 

\vspace{12pt}
\end{document}